\documentclass[11pt,a4paper]{article}
\usepackage{authblk}
\usepackage{stackengine}
\usepackage[hyperref]{emnlp2018}
\usepackage{times}
\usepackage{latexsym}

\usepackage{url}
\usepackage{xcolor}
\usepackage{upgreek}
\usepackage[shortlabels]{enumitem}
\usepackage{tabu}
\setlist[enumerate]{nosep}
\usepackage{floatrow}
\usepackage{multirow}
\usepackage{makecell}
\usepackage{amsmath}

\usepackage{booktabs}
\usepackage{graphicx}
\usepackage{amsfonts}
\usepackage{tabularx}
\usepackage{bm}
\aclfinalcopy

\usepackage{graphicx}
\graphicspath{ {./images/} }
\usepackage[ruled]{algorithm2e}
\usepackage{amsthm}
\usepackage{amssymb}

\makeatletter
\newcommand{\vo}{\vec{o}\@ifnextchar{^}{\,}{}}
\makeatother
\DeclareFloatFont{tiny}{\tiny}
\usepackage{silence}
\usepackage{tabularx}
\usepackage{pifont}
\usepackage{subcaption}

\title{CHOLAN: A Modular Approach for Neural Entity Linking on Wikipedia and Wikidata}

\author[1]{Manoj Prabhakar Kannan Ravi}
\author[3]{Kuldeep Singh}
\author[2]{Isaiah Onando Mulang'}
\author[4]{\\Saeedeh Shekarpour}
\author[5]{Johannes Hoffart}
\author[2]{Jens Lehmann}

\affil[1]{\stackunder{Hasso Plattner Institute, University of Potsdam, Potsdam, Germany}{\ttfamily manoj.prabhakar@hpi.de}\vspace{0.3em}}
\affil[2]{\stackunder{Smart Data Analytics, University of Bonn, Bonn, Germany}{\ttfamily \{mulang,jens.lehmann\}@cs.uni-bonn.de}\vspace{0.3em}}
\affil[3]{\stackunder{Zerotha Research and Cerence GmbH, Aachen, Germany}{\ttfamily kuldeep.singh1@cerence.com}\vspace{0.3em}}
\affil[4]{\stackunder{University of Dayton, Dayton, USA}{\ttfamily sshekarpour1@udayton.edu}\vspace{0.3em}}
\affil[5]{\stackunder{Goldman Sachs, Frankfurt, Germany}{\ttfamily johannes.hoffart@gs.com}\vspace{0.3em}}

\date{}

\begin{document}
\setlength{\abovedisplayskip}{3pt}
\setlength{\belowdisplayskip}{3pt}

\maketitle
\vspace*{5em}

\begin{abstract}
In this paper, we propose CHOLAN, a modular approach to target end-to-end entity linking (EL) over knowledge bases. CHOLAN consists of a pipeline of two transformer-based models integrated sequentially to accomplish the EL task. The first transformer model identifies surface forms (entity mentions) in a given text. For each mention, a second transformer model is employed to classify the target entity among a predefined candidates list. The latter transformer is fed by an enriched context captured from the sentence (i.e. local context),  and entity description gained from Wikipedia. Such external contexts have not been used in state of the art EL approaches. Our empirical study was conducted on two well-known knowledge bases (i.e., Wikidata and Wikipedia). The empirical results suggest that CHOLAN outperforms state-of-the-art approaches on standard datasets such as CoNLL-AIDA, MSNBC, AQUAINT, ACE2004, and T-REx. 
\end{abstract}


\section{Introduction} \label{sec:introduction}

The explicit schema, graph-based structure, and interlinking nature of information represented in publicly available knowledge graphs (KGs) e.g., DBpedia \cite{DBLP:conf/semweb/AuerBKLCI07}, Freebase \cite{DBLP:conf/aaai/BollackerCT07},  Wikidata \cite{DBLP:conf/www/Vrandecic12} or knowledge bases (KBs) such as Wikipedia; introduce a new landscape of features, as well as structured knowledge and embeddings. Researchers have developed several techniques to align information available in unstructured text to the concepts of these KGs \cite{DBLP:conf/aaai/WuFZ19,broscheit2019investigating}.

\paragraph{}
\vspace*{4em}
\noindent End-to-end Entity Linking (hereafter EL) task follows this direction; such that, given a sentence EL first identifies the entity mention in the sentence, then maps these mentions to the most likely KG/KB entities. 
The EL comprises of a three-step process. With respect to the given example sentence \textit{Soccer: Late Goals Give Japan win Over Syria}, the first step called mention detection (MD) identifies the surface forms \textit{Japan} and \textit{Syria}. The next step is candidate generation (CG) aiming to find a list of possible entity candidates in the KG/KB for each entity mention. 
For example, the candidates list for entity mention \textit{Japan} consists in part of \textit{Japan national football team, Japan (country), Japan (Band)} and for \textit{Syria} is \textit{Syria (Roman province), Syria national football team, Greater Syria}. Finally, the third step deals with the entity disambiguation (ED) which employs the co-reference and contextual features to discriminate the most likely entity from the candidates list e.g., \textit{Japan national football team} and \textit{Syria national football team} are correct entities.

Entity Linking approaches are broadly categorised into three categories. The initial attempts  \cite{DBLP:conf/emnlp/HoffartYBFPSTTW11,piccinno2014tagme} solve MD and ED as independent sub-tasks of EL (i.e., a pipeline based system). However, these approaches exhibit a behaviour where errors propagate from MD to ED hence might downgrade the overall performance of the system. The second category has emerged in an attempt to mitigate these errors, where researchers focused on jointly modelling MD and ED, emphasising the importance of the mutual dependency of the two sub-tasks \cite{kolitsas2018end}. These two EL approaches depend on an intermediate candidate generation step and rely on a pre-computed list of entity candidates. For example, \cite{kolitsas2018end} propose a joint MD and ED model and inherits the candidate list from \cite{ganea2017deep}. The third approach combines the three sub-steps in a joint model and illustrates that each of those tasks is interdependent \cite{durrett2014joint,broscheit2019investigating}. 

The recent EL approaches focus on jointly modelling two or three subtasks \cite{sevgili2020neural}. Furthermore, the NLP research community has extensively used transformers in end-to-end models for entity linking (\citealt{broscheit2019investigating}, \citealt{peters2019knowledge}, and \citealt{evry2020empirical}). Nevertheless, these works report less performance than \cite{kolitsas2018end}, which is a bi-LSTM based model. The observations regarding the limited performance of transformer-based models for the EL motivate our work, and in this paper, our focus is to understand the bottlenecks in the entity linking process. We argue that the less studied task in literature, i.e., candidate generation, has an essential role in the EL models' performance, which has not been a focus in the recently proposed transformer-based entity linking models. 

\begin{figure*}
	\centering
	\includegraphics[width=\textwidth]{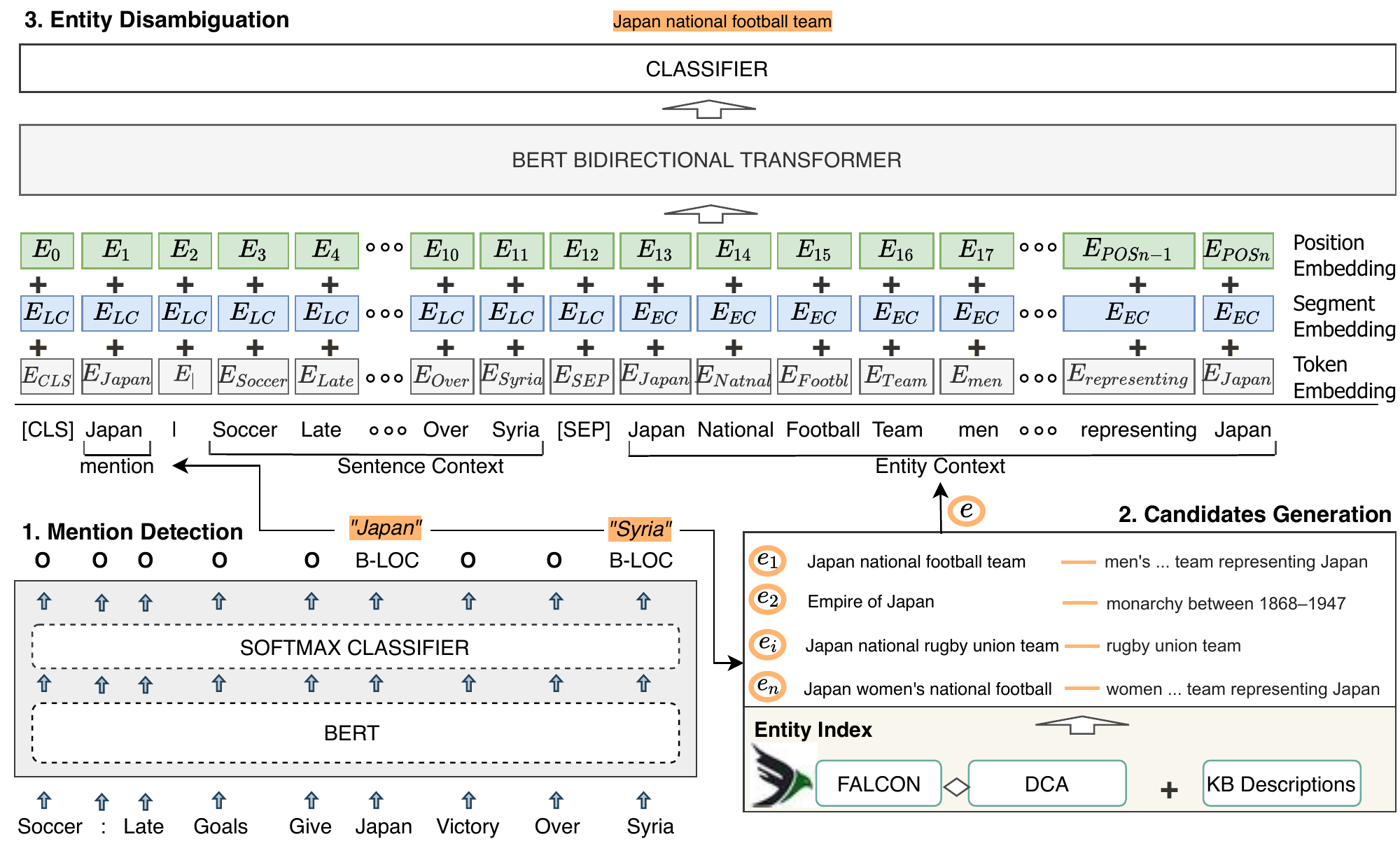}
	\caption{CHOLAN has three building blocks: i) BERT-based Mention Detection that identifies entity mentions in the text ii) Candidate Generation that retrieves a set of entities for the mention  iii) Entity Disambiguation: employs BERT transformer model powered by background knowledge from KB and local sentential context.}
	\label{fig:cholan-approach}
	    \vspace{-2mm}
	\end{figure*}

In this paper, we \textbf{hypothesise} that the transformer models, though trained on a large corpus, may require additional task-specific contexts.
Furthermore, inducing the context at the entity disambiguation step may positively impact the overall performance, which has not been utilised in the state of the art methods due to monolithic implementations \cite{kolitsas2018end,peters2019knowledge,broscheit2019investigating,evry2020empirical}. Subsequently, we deviate from the joint modelling of two or three subtasks of the EL and revert to the methodology opted by earlier EL systems in 2011 \cite{DBLP:conf/emnlp/HoffartYBFPSTTW11}, i.e. treat each sub-task independently. As such, we study the research question: \textbf{RQ}: \textit{what is the impact of each sub-task (aka component) on the overall outcome of the transformer-based entity linking approach?} We propose an intuitive novel approach named CHOLAN, comprising a modular architecture of two transformer models to solve MD and ED independently. In the first step, CHOLAN employs BERT \cite{devlin2019bert} model to identify mentions of the entities in an input sentence. The second step involves expanding each mention with a list of KB entity candidates. Finally, the entity mention, sentence (local context), an entity candidate, and entity Wikipedia description (entity context) are fed as input sequences in the second BERT based model to predict the correct KB entity (cf. Figure \ref{fig:cholan-approach}). We train MD and ED steps independently during training, and while testing, we run the CHOLAN pipeline end-to-end for predicting the KB entity. The following are the novel features of CHOLAN:
\noindent\begin{itemize}
\itemsep-.4em 
\item The core focus of the approach is to flexibly induce \underline{external context} and \underline{candidate lists} in a transformer-based model to improve the EL performance.
CHOLAN is independent of a particular candidate list and additional background context. We study four different configurations of CHOLAN to demonstrate the impact of candidate generation step and background knowledge (i.e. entity and sentential context) induced in the model. CHOLAN achieves a new state of the art performance on several datasets: T-REx \cite{DBLP:conf/lrec/ElSaharVRGHLS18} for Wikidata; AIDA-B, MSBC, AQUAINT, and ACE2004 for Wikipedia \cite{DBLP:conf/emnlp/HoffartYBFPSTTW11,guo2018robust}. 
\item CHOLAN is the first approach which is empirically demonstrated to be transferable across KBs having completely different underlying structure and schema i.e., on semi-structured Wikipedia and fully structured Wikidata. 
\end{itemize} 
The implementation is publicly available\footnote{\url{https://github.com/ManojPrabhakar/CHOLAN}}. 
The paper is structured as follows: 
next section summarises the related work. Section \ref{sec:problem} describes the problem statement and approach. Section \ref{sec:experiment} explains the experimental settings followed by results in \ref{sec:results}. We conclude in Section \ref{sec:conclusion}.\\

\section{Related Work} \label{sec:related}
\textbf{Mention Detection (MD)}: The first attempt to organise a named entity recognition (NER) task traced back to 1996 \cite{grishman1996message}. Since then, numerous attempts have been made ranging from conditional random fields (CRFs) with features constructed from dictionaries \cite{rocktaschel2013wbi} or feature-inferring neural networks \cite{collobert2008unified}. Recently, contextual embedding based models achieve state of the art for NER/MD task \cite{akbik-etal-2018-contextual,devlin2019bert}. We point to the survey by \citet{yadav2018survey} for details about NER. Few early EL models have performed MD task independently \cite{ceccarelli2013dexter,cornolti2016piggyback}. \\
\textbf{Candidate Generation (CG)}: There are four prominent approaches for candidate generation. First is a direct matching of entity mentions with a pre-computed candidate set \cite{zwicklbauer2016robust}. The second approach is the dictionary lookup, where a dictionary of the associated aliases of entity mentions is compiled from several knowledge base sources (e.g. Wikipedia, Wordnet) \cite{sevgili2020neural,fang2019joint,cao2017bridge}. The third approach is to generate entity candidates using empirical probabilistic entity-map $p(e|m)$. The $p(e|m)$ is a pre-calculated prior probability of correspondence between positive mentions and entities. A widely used entity map was built by \cite{ganea2017deep} from Wikipedia hyperlinks, Crosswikis \cite{spitkovsky2012cross} and YAGO \cite{DBLP:conf/emnlp/HoffartYBFPSTTW11} dictionaries. End-to-end EL approaches such as \cite{kolitsas2018end,cao2018neural} relies on the entity map built by Ganea and Hofmann. The next approach for generating the candidates is proposed by \cite{DBLP:conf/naacl/SakorMSSV0A19}. Authors build a local KG by expanding entity mentions using Wikidata and DBpedia entity labels and associated aliases. The local KG can be queried using BM25 ranking algorithm \cite{logeswaran2019zero}. The modular architecture of CHOLAN gives us the flexibility to experiment with several ways of generating entity candidates. Hence, we reused candidate list proposed by \cite{ganea2017deep} and built a new CG approach based on \cite{DBLP:conf/naacl/SakorMSSV0A19}.\\
\textbf{End to End EL:} Few EL approaches accomplish MD and ED tasks jointly. \cite{nguyen2016j} propose joint recognition and disambiguation of named-entity mentions
using a graphical model and show that it improves EL. 
The work in \cite{kolitsas2018end} also proposes a joint model for MD and ED. Authors use a bi-LSTM based model for mention detection and computes the similarity between the entity mention embedding and set of predefined entity candidates.
The work in \cite{broscheit2019investigating} employs BERT to jointly model three subtasks of the EL. Author employ an entity vocabulary of 700K top most frequent entities to train the model.
Work in \cite{evry2020empirical} uses a Transformer architecture with large scale pre-training from Wikipedia links for EL. For CG, authors train the model to predict BIO-tagged mention boundaries to disambiguate among all entities. For Wikidata KG, Opentapioca is an entity linking approach which relies on a heuristic-based model for disambiguation of the mentions in a text to the Wikidata entities \cite{delpeuch2019opentapioca}. Arjun \cite{Mulang2019ContextawareEL} is the most similar to our approach CHOLAN and trains two independent neural models for MD and ED. It generates candidates on the fly using a Wikidata entity alias map. Arjun does not induce any context in the model. 
%
%
\section{Problem Statement and Approach}

\label{sec:problem}
We formally define EL task as follows: given an input sequence of words W = \{$w_1, w_2, w_3, …, w_n$\}, and a 
set of entities denoted by $\mathcal{E}$ from a KG/KB. The EL task aligns the text into a subset of entities represented as $ \Theta: W \rightarrow  \mathcal{E'}$ where $\mathcal{E'} \subset \mathcal{E}$. We formulate the EL task as a three step process in which the first step is the mention detection (MD).  
The MD is a function $\theta_1: W \rightarrow  \mathcal{M}$, where the set of mentions is denoted by $\mathcal{M}=(m_1,m_2,...,m_k)$ ($k\leq n$) and each mention $m_{x}$ is a sequence of words starting from $i$ to end position $j$: $m_x^{(i,j)}=(w_i,w_{i+1},...,w_j)$ ($0<i,j\leq n$). 
The next task is candidate generation where for each mention  $m_{x}$ a set of candidates $C(m_x)$= \{$e_1^{x},...,e_n^{x}|e_i^{x} \in \mathcal{E}\}$ is derived.
Finally, the entity disambiguation (ED) task aims to map each mention  $m_x \in \mathcal{M}$ to the most likely entity from its list of candidates.
In our case, we model the ED task as a classification task and augment the input with extra signals as context. For every candidate entity $c_i \in C(m_x)$, the model estimates a probability $p_i$, thus the most likely entity is the one with the highest probability as $\gamma = \arg\max_{p_i}  \{  \mathcal{P}(p_i\mid m_x,c_i^x, W, C) \} $ where $W$ and $C$ are the input representations respectively for the given sentence (local context) and the context derived from KG/KB. As such the probability of score $p_i$ is conditioned not only on $m_x$ and $c_i^x$ but also on $W$ and $C$ as contextual parameters.


\subsection{CHOLAN Approach} \label{sec:cholan}
The CHOLAN architecture comprises of three main modules as illustrated in Figure \ref{fig:cholan-approach}. 
\subsubsection{Mention Detection (MD)} \label{sec:mention}
We adapt the vanilla BERT~\cite{devlin2019bert} model for the task of entity mention detection in an unstructured text. For each input sentence, we append the special tokens [CLS] and [SEP] to the beginning and end of the sentence, respectively. This is then used as input to the model which learns a representation of the tokens in the sentence. We then introduce a (logistic regression based) classification layer on top of the BERT model to determine named entity tags for each token following the BIO format~\cite{DBLP:journals/corr/cs-CL-0306050}. Our BERT$^\dagger$ model is initialised using publicly available weights from the pretrained BERT$_{BASE}$ model and is fine-tuned to the specific dataset for detecting a mention $m_{i}$. Please note that BERT$_{BASE}$ model is the latest approach which successfully outperformed in various NLP tasks, including MD. Thus, we reuse this model for the completion of our approach. 
\begin{align}
\label{eq:NER}
m_i = BERT^\dagger(w_i)
\end{align}
\subsubsection{Candidate Generation (CG)} \label{sec:candidate}
One of the critical focus of CHOLAN is to understand the bottleneck at the CG step. Hence, we reuse the DCA candidate list and propose a novel candidate list to understand the candidate generation impact on overall EL performance.

\textbf{DCA Candidates}: \cite{yang2019learning} adapts the probabilistic entity-map $p(e|m)$ created by \cite{ganea2017deep} (cf. section \ref{sec:related}) to calculate the prior probabilities 
of candidate entities for a given mention. In the probabilistic entity-map, each entity mention has 30 potential entity candidates.
Yang and colleagues also provide associated Wikipedia description of each entity. In CHOLAN, we \textit{reuse} candidate set $C(m)$ provided by \cite{yang2019learning} and further consider associated Wikipedia entity descriptions.\\
\textbf{Falcon Candidates}: \cite{DBLP:conf/naacl/SakorMSSV0A19} created a local index of KG items from Wikidata entities expanded with entity aliases. For example, in Wikidata the entity Q33\footnote{\url{https://www.wikidata.org/wiki/Q33}} has the label "Finland". Sakor and colleagues expanded the entity label with other aliases from Wikidata such as ``Finlande", ``Finnia", ``Land of Thousand Lakes", ``Suomi", and ``Suomen tasavalta". We adopt this local KG index to generate entity candidates per entity mention in the employed datasets.  The local KG has a querying mechanism using BM25$^\dagger$ algorithm (cf. equation \eqref{eq:CG}) and ranked by the calculated score. We build a predefined candidate set using the top 30 Wikidata entity candidates in $C\_Falcon(m)$ for each entity mention. 
We enrich the candidates set obtained from Wikidata by the correspondence from Wikipedia.
We also add the first paragraph of Wikipedia as entity descriptions (only if Wikidata entity has corresponding Wikipedia page) to the hyperlinks. By selecting two different candidate list, our idea is to understand the impact of candidate generation step on end-to-end entity linking performance.
\begin{align}
\label{eq:CG}
    e_i = BM25^\dagger(m_i)
\end{align}

\subsubsection{Entity Disambiguation (ED)} \label{sec:entity}
In order to use the power of the transformers, we propose ``WikiBERT" to perform the ED task. In WikiBERT, our novel methodological contribution is the induction of local sentential context and global entity context at the ED step in a transformer model, which has not been used in the recent EL models. WikiBERT is derived from the vanilla BERT$_{BASE}$ model and fine-tuned on the two EL datasets (CoNLL-AIDA and T-REx). 
We view the ED task as sequence classification task. The input to our model is a combination of two sequences. The first sequence $S_1$ concatenates the entity mention $m \in \mathcal{M}$ and sentence $\mathcal{W}$ where the sentence acts as a \underline{local context}. The second sequence $S_2$ is a concatenation of entity candidate $e \in C(m)/C\_Falcon(m)$(obtained from Equation \ref{eq:CG}) and its corresponding Wikipedia description (\underline{entity context} $ct_i$). The two sequences are paired together with special start and separator tokens: ([CLS] $S_1$ [SEP] $S_2$ [SEP]). The sequences are fed into the model which in turn learns the input representations according to the architecture of BERT \cite{devlin2019bert}. Any given token (local context word, entity mention, or entity context words) is a summation of the three embeddings :
\begin{enumerate}[i.]
    \item \textit{Token embedding}: refers to the embedding of the corresponding token. We make note here on specific tokens that comprises the input representations for our model more specialised as compared to other fine-tuning tasks. The entity mention tokens appended at the beginning of $S_1$ and separated from the sentence context tokens by a single vertical token bar $\vert$, likewise, for the entity context sequence $S_2$, we prepend the entity title tokens from the KB before adding the descriptions.
    \item \textit{Segment embedding}: each of the sequences receive a single representation such that the segment embedding for the local context $E_{LC}$ refers to the representation for $S_1$ whereas $E_{EC}$ is the representation of $S_2$
    
    \item \textit{Position embedding}: represents the position of the token in an input sequence. A token appearing at the i-th position in the input sequence is represented with $E_i$
\end{enumerate} 
To train the model, we use the negative sampling approach similar to \citet{yamada2019pre}. The candidate list is generated for each identified mention.
The desired entity candidate item is labelled as one, and the rest of the incorrect candidate items (from candidate list) are labelled as zero for a given mention. This process iterates over all the identified mentions using Equation \ref{eq:NER}.

The training process fine-tunes BERT using the contextual input from sentence and Wikipedia resulting into the WikiBERT model (Equation \eqref{eq:NED}). The model predicts the relatedness of the two sequences by classifying it as either positive or negative.
\begin{align}
\label{eq:NED}
    e_i = WikiBERT(m_i, e_i, ct_i)
\end{align}

\section{Experimental Setup} \label{sec:experiment}
\subsection{Datasets}
For Wikidata EL, we rely on T-REx dataset \cite{DBLP:conf/lrec/ElSaharVRGHLS18}.
We adapt the subset of T-REx used by \citet{Mulang2019ContextawareEL} for a fair evaluation setting. The dataset contains 983,257 sentences (786,605 in training and 196,652 in the test set) accommodating 3,133,778 instances of surface forms which are linked to 85,628 distinct Wikidata entities. T-REx does not have a separate validation set to fine-tune the hyperparameters. Therefore, we further divide the train set into a 90:10 ratio for training and validation.

For EL over Wikipedia, we adapt standard dataset CoNLL-AIDA proposed by \cite{DBLP:conf/emnlp/HoffartYBFPSTTW11} for the training. The dataset contains 18,448 linked mentions in 946 documents, a test set of 4,485 mentions in 231 documents, and a validation set of 4,791 mentions in 216 documents. For testing, we use AIDA-B (test) dataset from \cite{DBLP:conf/emnlp/HoffartYBFPSTTW11} and  MSNBC, AQUAINT, ACE2004 datasets from \cite{guo2018robust}. 
\subsection{Models for Comparison} 
\subsubsection{Baselines over Wikidata}
We now briefly explain Wikidata baselines.\\
\textbf{1.} OpenTapioca \cite{delpeuch2019opentapioca}: is a heuristic-based end-to-end approach that depends on topic similarity and mapping coherence for linking Wikidata entity in an input text. \\
\textbf{2.} Arjun \cite{Mulang2019ContextawareEL}: is a pipeline of two attentive neural networks employed for MD and ED. Arjun is the SotA, and we take baseline values from Arjun's paper. 

\subsubsection{Baselines over Wikipedia}
\textbf{1.} \cite{DBLP:conf/emnlp/HoffartYBFPSTTW11}: build a weighted graph of entity mentions and candidate entities. Then, the model computes a dense subgraph that predicts the best joint mention-entity mapping.\\
\textbf{2.} DBpedia Spotlight \cite{MendesJGB11} proposes a probabilistic model and relies on the context of the text to link the entities.\\
\textbf{3.} KEA \cite{steinmetz2013semantic} employs a linguistic pipeline coupled with metadata generated from several Web sources. The candidates are ranked using a heuristic approach.\\
\textbf{4.} Babelfy \cite{moro2014entity} is a graph-based approach that uses loose identification of candidate meanings coupled with the densest subgraph heuristic to link the entities.\\
\textbf{5.} \citet{piccinno2014tagme}: to solve entity linking, authors focus on mentions recognition and annotations pruning to propose a voting algorithm for entity candidates using PageRank.\\
\textbf{6.} \citet{kolitsas2018end} train MD and ED task jointly using word and character-level embeddings. The model reuses candidate set from \cite{ganea2017deep} and generates a global voting score to rank the entity candidates.\\
\textbf{7.} \citet{peters2019knowledge} induce multiple KBs into a large pretrained BERT model with a knowledge attention mechanism.\\
\textbf{8.} \citet{broscheit2019investigating} trains MD, CG, ED task jointly using a BERT-based model. Besides, an entity vocabulary containing 700K most frequent entities in English Wikipedia was utilised. \\
\textbf{9.} \citet{evry2020empirical} consider large scale pretraining from Wikipedia links as the context for a transformer model to predict KB entities.\\
In Wikipedia-based experiments, we report values from \cite{evry2020empirical} and \cite{kolitsas2018end} for AIDA-B test set. 
On MSNBC (MSB), AQUAINT (AQ), and ACE2004 (ACE) test datasets, only \cite{kolitsas2018end}, DBpedia Spotlight \cite{MendesJGB11}, KEA \cite{steinmetz2013semantic}, and Babelfy \cite{moro2014entity} report the values and we compare against them. 
\begin{table}[!htp]
\centering
\begin{tabular}{ll}
\toprule
{\textbf{Hyper-parameters}} &  {\textbf{Value}}  \\
\midrule
Epochs & 4 \\
Batch size & 8 \\
Learning rate & $2e^{-5}$ \\
Learning rate decay & linear \\
Adam $\beta_{1}$ & 0.9  \\
Adam $\beta_{2}$ & 0.999  \\
dropout & 0.1 \\
Loss Function & Cross-Entropy \\
Classifier & Softmax \\
\bottomrule
\end{tabular}
\caption{Hyper-parameters during fine-tuning.}
\label{tab:Hyperparameters}
\end{table}

\subsection{CHOLAN Configurations} \label{sec:config}
We configure CHOLAN model applying various candidate generation approaches detailed below.\\
\textbf{CHOLAN-Wikidata}: we train the model using T-REx dataset and employ $C\_Falcon(m)$ candidate set. The ED model (WikiBERT) is fed with the sentential context but not with entity description as not all Wikidata entities have a corresponding Wikipedia entity. \\
\textbf{CHOLAN-Wiki+FC}: is trained on CoNLL-AIDA \cite{DBLP:conf/emnlp/HoffartYBFPSTTW11}. For CG step, we employ Falcon candidate set $C\_Falcon(m)$. 
Here, the ED model (WikiBERT) is only fed with the sentential context.\\
\textbf{CHOLAN-Wiki+DCA}: We train the MD and ED models on CoNLL-AIDA. The CG step involves DCA candidate set $C(m)$. During ED step (WikiBERT), Wikipedia descriptions associated with each entity is fed along with sentential context.\\
\textbf{CHOLAN}: inherits \textbf{CHOLAN-Wiki+FC} but in addition, Wikipedia entity description is induced into the ED model (WikiBERT).
\subsection{Metrics and Hyper-parameters}
On Wikidata-based experiments, we employ standard metrics of accuracy i.e., precision (P), recall (R), and F-score (F) same as \cite{Mulang2019ContextawareEL}. For Wikipedia-based datasets, we use Micro-F1 score in strong matching setting \cite{kolitsas2018end}. The strong matching needs exactly predicting the gold mention (i.e. target entity mention) 
boundaries and its corresponding entity annotation in the KB. To compare the recalls of two CG approaches, we report the performance on gold recall. Gold recall is the percentage of entity mentions for which the candidate set contain the ground truth entity \cite{yao2019kg}. \\
We have implemented all our models in PyTorch\footnote{\url{https://pytorch.org/}} and optimized using Adam~\cite{DBLP:journals/corr/KingmaB14}. We used the pre-trained BERT models from the Transformers library \cite{Wolf2019HuggingFacesTS}. We ran all the experiments on a single GeForce GTX 1080 Ti GPU with 11GB size.
Table \ref{tab:Hyperparameters} outlines the hyper-parameters used in the fine-tuning on both the datasets. We followed the standard settings suggested by \cite{devlin2019bert}. The average run time is 9.31 hours/epoch for CHOLAN and without description, it was 7.23 hours/epoch.




\section{Results}\label{sec:results}
We study the following research question:\textit{what is the impact of each sub-task (aka component) on the overall outcome of the transformer-based entity linking approach?} We further investigate a sub-research question: how do the external context and the candidate generation step impact the overall performance of CHOLAN?
Our every experiment systematically studies the research questions in different settings. 

\begin{table}[!htp]
    \centering
    \begin{tabular}{p{3cm}p{0.65cm}p{0.7cm}p{0.65cm} }
     
     \toprule
     \textbf{Model} & \textbf{P} & \textbf{R} & \textbf{F}\\
    \midrule
      \citealt{delpeuch2019opentapioca}  & 40.7    &\textbf{82.9} &57.9\\
      \citealt{Mulang2019ContextawareEL}  & 71.4    &71.2 &71.3\\
     \midrule
     CHOLAN-Wikidata &  \textbf{75}  & 76 &\textbf{75.4}\\
    \bottomrule
    \end{tabular}
    \caption{Comparison on T-REx test set for Wikidata EL. Best values in bold.}
    \label{tab:trex}
    \vspace{-2mm}
\end{table}

\subsection{Results on Wikidata dataset}
Table \ref{tab:trex} summarises CHOLAN performance on T-REx dataset. CHOLAN-Wikidata configuration outperforms the baselines. We dig deeper into our reported values. We observe that for MD task, our F-score is 94.3 (compared to 77 F-score of Arjun \cite{Mulang2019ContextawareEL}). However, the gold recall for CG step is 81.2. We generate the entity candidates using an information retrieval approach (BM25$^\dagger$ algorithm) to get the top 30 candidates based on the confidence score. The Wikidata KG is challenging, and many labels share the same name. It contributes to a large loss in the F-score for the CG step. For instance, the entity mention ``National Highway'' matches exactly with four Wikidata ID labels while 2,055 other entities contain the full mention in their labels. Please note that we did not perform retraining of \cite{kolitsas2018end} (SOTA on Wikipedia EL) on the T-REx dataset since we determined that the model is tightly coupled and relies on pre-computed Wikipedia candidate list from \cite{ganea2017deep}. 

\subsubsection{Ablation Study on Wikidata}
We study the impact of local context on the performance of CHOLAN. Therefore, we exclude the sentence as input in the ED step at training and testing time. Hence, the inputs to the ED model are only entity mention and the entity candidates gained from the CG step. We observe that the performance drops when the local sentential context is not fed (cf. Table \ref{tab:trexcontext}). It justifies our choice to feed the model by the sentence during the ED task. 

\begin{table}[!htp]
    \centering
    \begin{tabular}{p{3.70cm}p{0.6cm}p{0.60cm}p{0.60cm} }
     
  \toprule
     \textbf{Model} & \textbf{P} & \textbf{R} & \textbf{F}\\
    \midrule
     CHOLAN-Wikidata &  \textbf{75}  & \textbf{76} &\textbf{75.4}\\
     CHOLAN-Wikidata (WLC$^\dagger$) & 72 & 73.5  & 72.7\\
   \bottomrule
    \end{tabular}
    \caption{The ablation study on T-REx test set for Wikidata EL. Best values in bold. WLC$^\dagger$ denotes model without local context. When the local sentential context is excluded from ED, the performance drops.}
    \label{tab:trexcontext}
    \vspace{-2mm}
\end{table}

\subsection{Results on Wikipedia datasets} \label{sec:wikipediadata}
Table \ref{tab:aidaB} reports the performance of CHOLAN's configurations on AIDA-B test set. 
The first configuration is "CHOLAN-Wiki+ FC" in which MD and ED models are trained using CoNLL-AIDA. We notice a clear jump in the performance. We then replaced the Falcon candidate list $C\_Falcon(m)$ with DCA candidates $C(m)$ resulting into "CHOLAN-Wiki+ DCA". In DCA candidates, the description of entities is attached. The performance is increased when an additional background knowledge as an entity description is fed. Our next configuration is CHOLAN where we attached Wikipedia entity descriptions in Falcon candidate list $C\_Falcon(m)$ (as a modification of "CHOLAN-Wiki+ FC"). This setting outperforms all the existing baselines and previous CHOLAN configurations. Our experiments illustrate the impact of CG step and background knowledge on end-to-end EL performance. The improvement of CHOLAN continues to the other three test datasets where the jump is significantly higher compared to the baselines (cf. Table \ref{tab:msb}). Reported values in Table \ref{tab:msb} also approves transferability of CHOLAN when we apply cross-domain experiments.

\begin{table}[ht!]
    \centering
    \begin{tabular}{p{4cm}p{1.6cm}}
    \toprule
     \textbf{Model} & \textbf{Micro F1}\\
   \midrule
     \citealt{DBLP:conf/emnlp/HoffartYBFPSTTW11}   &72.8\\
      \citealt{MendesJGB11}   &57.8\\
      \citealt{steinmetz2013semantic}   &42.3\\
      \citealt{moro2014entity}  &48.5\\
     \citealt{piccinno2014tagme} & 73 \\
     \citealt{kolitsas2018end}& \underline{82.4} \\
     \citealt{peters2019knowledge} & 73.7\\
     \citealt{broscheit2019investigating} & 79.3 \\
     \citealt{evry2020empirical} & 76.7 \\
    \midrule
     CHOLAN-Wiki+ FC & 75.1\\
     CHOLAN-Wiki+ DCA&  77.5\\
     CHOLAN & \textbf{83.1} \\
     \bottomrule
    
    \end{tabular}
\caption{Comparison on \textit{AIDA-B}. Best value in bold and previous SOTA value is underlined.}
\label{tab:aidaB}
\vspace{-2mm}
\end{table}
\begin{table}[!htp]
    \centering
    \begin{tabular}{p{3.5cm}p{0.65cm}p{0.7cm}p{0.65cm} }
    \toprule
     \textbf{Model} & \textbf{MSB} & \textbf{AQ} & \textbf{ACE}\\
   \midrule
      \citealt{MendesJGB11}  & 40.6    &\underline{45.2} &60.5\\
      \citealt{steinmetz2013semantic}  & 30.9    &35.9 &40.3\\
      \citealt{moro2014entity}  & 39.7   &35.8 &17.8\\
      \citealt{kolitsas2018end}  & \underline{72.4} &40.4 & \underline{68.3}\\
    \midrule
    CHOLAN-Wiki+ FC  &   77.8  & 70 &85.7\\
     CHOLAN-Wiki+ DCA &   78.3 & 75.9 &71.3\\
      CHOLAN  &  \textbf{83.4} & \textbf{76.8} &\textbf{86.8}\\
   \bottomrule
    \end{tabular}
    \caption{The micro F1 scores are listed from the comparative study over three datasets (out of domain). The model is trained over CoNLL-AIDA dataset. Best value in bold and previous SOTA value is underlined.}
    \label{tab:msb}
    \vspace{-2mm}
\end{table}

\subsubsection{Ablation Study on Wikipedia}
We conducted three ablation studies to understand the behaviour of CHOLAN's configurations over Wikipedia datasets. The first study is to calculate the Gold recall values for various datasets. CHOLAN uses the candidates from $C\_Falcon(m)$ candidate set for each entity mention. While generating the candidate set from local KG of \cite{DBLP:conf/naacl/SakorMSSV0A19}
we observe a drop in the Gold recall as reported in Table \ref{tab:recall}. 
CG plays a crucial role in trading off precision and recall. We conclude that more robust CG approaches likely impact overall performance. The second ablation study is about to calculate the performance of our configurations for ED step, i.e., running WikiBERT in isolation. Here, we assume that all entities are truly recognised; thus, our focus of the study is the ED model. We report the impact of various candidate generation approaches on the ED model in Table \ref{tab:aidaED}. The significant jump in the performance from "CHOLAN-Wiki+FC Vs CHOLAN" contributes to the additional background knowledge provided in CHOLAN as entity candidate descriptions. 
The third ablation study tests the impact of sentential context fed into two configurations on a Wikipedia dataset. Table \ref{tab:aidacontext} reports the achieved performance after excluding sentence as the additional context. Obviously, the performance decreases. The model shows similar behaviour on T-REx in Table \ref{tab:trexcontext}. These observations confirm our hypothesis as the ED model is enhanced using additional contexts.

\begin{table}[!htp]
    \centering
    \begin{tabular}{p{2.6cm}p{1.2cm}p{0.6cm}p{0.6cm}p{0.6cm} }
    \toprule
     \textbf{Model} & \textbf{AIDA-B} & \textbf{MSB} & \textbf{AQ} & \textbf{ACE}\\
    \midrule
      Falcon Candidates &94& 93.8  &85.3 &97.3\\
      DCA  Candidates & 98.3&  98.5 & 94.2 &90.6\\
    \bottomrule
    \end{tabular}
    \caption{Gold Recall for Candidate Generation techniques over Wikipedia test datasets.}
    \label{tab:recall}
    \vspace{-2mm}
\end{table}

\begin{table}[ht!]
    \centering
    \begin{tabular}{p{3.7cm}p{1.4cm}}
   \toprule
     \textbf{Model} & \textbf{Micro F1}\\
    \midrule
     \citealt{kolitsas2018end}& \underline{83.8} \\
     \midrule
     CHOLAN-Wiki+ FC & 78.4\\
     CHOLAN-Wiki+ DCA& 79.1\\
     CHOLAN & \textbf{85.7} \\
    \bottomrule
    \end{tabular}
\caption{Comparison on \textit{AIDA-B} for ED. Best score in bold and previous SOTA value is underlined.}
\label{tab:aidaED}
\vspace{-2mm}
\end{table}

\begin{table}[ht!]
    \centering
    \begin{tabular}{p{4.5cm}p{1.7cm}}
   \toprule
     \textbf{Model} & \textbf{Micro F1}\\
    \midrule
    CHOLAN-Wiki+ DCA & 77.5\\
    CHOLAN-Wiki+ DCA (WLC$^\dagger$) &71.2\\
    CHOLAN  &  \textbf{83.1} \\
    CHOLAN (WLC$^\dagger$) & 79.6 \\
    \bottomrule
    \end{tabular}
\caption{Ablation study on \textit{AIDA-B}. We observe that when local sentential context is removed from ED step, the performance drops. Best values in bold. WLC$^\dagger$ denotes model without local context.}
\label{tab:aidacontext}
\vspace{-2mm}
\end{table}

\section{Conclusions} \label{sec:conclusion}
In the last two years, the NLP research community has extensively tried transformer-based models for the EL task. However, the performance remained lower than \citet{kolitsas2018end}. This paper combines the traditional software engineering principle of modular architecture with the context-induced transformers to effectively solve the EL task. Our reason to deviate from an end-to-end architecture was to provide full flexibility to our system in terms of candidate generation list, underlying KG, and induction of the context at the ED step. We attribute CHOLAN's outperformance to the following reasons: 1) the modular architecture, which brings flexibility and interoperability as CHOLAN can treat each task independently. \citet{kolitsas2018end} reports that shifting towards joint modelling of MD and ED tasks helps mitigate error propagation from MD to ED. However, the performance of BERT$_{BASE}$ for the MD task is significantly high (92.3 on AIDA-B and 94.3 F1-score on T-REX calculated by us) remarkably reducing the errors in MD. CHOLAN leverages this capability in the MD subtask, placing more focus on CG and ED tasks. 2) The flexibility in architecture further permits us to induce sentence and entity descriptions as additional contexts. Furthermore, using candidate list in plug and play manner has resulted in a significant increase in the performance. In earlier transformer approaches, the implementation is monolithic and context is not utilised. 
There are scopes for improvement in our approach.
\citet{wu2019zero} introduces a novel CG method that retrieves candidates in a dense space defined by a bi-encoder and can be used as alternate CG approach. We aim for scaling CHOLAN to multilingual entity linking as a viable next step.

\bibliographystyle{acl_natbib_nourl}
\bibliography{emnlp2018} 

\end{document}